\documentstyle[AAAI]{article}

\input{psfig}

\newcommand{\datac}{\mathrm{DC}}

\newcommand{\at}{{\mathit At}}

\newtheorem{theorem}{Theorem}

\begin{document}

\title{{\tt dcs}: An Implementation of DATALOG with Constraints}

\author{Deborah East \and Miros\l aw Truszczy\'nski\\
		Department of Computer Science\\ 
		University of Kentucky\\ 
		Lexington KY 40506-0046, USA \\ 
		email: deast$|$mirek@cs.uky.edu} 
            
\maketitle

\begin{abstract}
\noindent 
Answer-set programming (ASP) has emerged recently as a 
viable programming paradigm.  
We describe here an ASP system, {\em DATALOG with constraints}
or $\datac$, based on non-monotonic logic. Informally, 
$\datac$ theories consist of propositional clauses
(constraints) and of Horn rules. The
semantics is a simple and natural extension of the semantics
of the propositional logic. However, thanks to the presence of Horn
rules
in the system, modeling of transitive closure becomes straightforward.
We describe the syntax, use and implementation of $\datac$ and
provide experimental results. 

\end{abstract}

\section{General Info}

$\datac$ is an answer set programming (ASP) system \cite{mt99} 
similar to propositional logic but extended 
to include Horn clauses.  The semantics of $\datac$ is a natural extension
of the semantics of propositional logic.
The $\datac$ system is implemented in two modules, {\tt ground} and {\tt dcs}, 
which are written in the 'C' programming
language  and compiled with {\tt gcc}. 
There are approximately 2500 lines of code for {\tt ground}
and approximately 1500 for {\tt dcs}. $\datac$ has been implemented on
both SUN SPARC and a PC running linux.  

A $\datac$ theory (or program) consists of constraints and Horn rules
(DATALOG program). This fact motivates our choice of terminology ---
DATALOG with constraints. We start a discussion of $\datac$ with the
propositional case. Our language
is determined by a set of atoms $\at$. We will assume that $\at$ is of the
form $\at = \at_C \cup \at_H$, where $\at_C$ and $\at_H$ are disjoint.

A $\datac$ theory (or program) is a triple $T_{dc}=(T_C,T_H,T_{PC})$, where\\
1. $T_C$ is a set of propositional clauses $\neg a_1\vee\ldots\vee \neg a_m
\vee b_1\vee\ldots\vee b_n$ such that all $a_i$ and $b_j$ are from
$\at_C$,\\
2. $T_H$ is a set of Horn rules $a_1\wedge\ldots\wedge a_m\rightarrow b$
such that $b\in \at_H$ and all $a_i$ are from $\at$,\\
3. $T_{PC}$ is a set of clauses over $\at$.\\

By $\at(T_{dc})$, $\at_C(T_{dc})$ and $\at_{PC}(T_{dc})$ we denote the set of atoms from
$\at$, $\at_C$ and $\at_{PC}$, respectively, that actually appear in $T_{dc}$.

With a $\datac$ theory $T_{dc} = (T_C,T_H,T_{PC})$ we associate a family of
subsets of $\at_C(T_{dc})$. We say that a set $M\subseteq \at_C(T_{dc})$ 
{\em satisfies} $T_{dc}$ (is an {\em answer set} of $T_{dc}$) if\\
1. $M$ satisfies all the clauses in $T_C$, and\\
2. the closure of $M$ under the Horn rules in $T_H$,
$M^c=LM(T_H\cup M)$ satisfies all clauses in $T_{PC}$ ($LM(P)$ denotes
the least model of a Horn program $P$).\\

Intuitively, the collection of clauses in $T_C$ can
be thought of as a representation of the constraints of the problem,
Horn rules in $T_H$ can be viewed as a mechanism to compute closures of sets
of atoms satisfying the constraints in $T_C$, and the clauses in $T_{PC}$ can
be regarded as constraints on closed sets (we refer to them as
{\em post-constraints}). A set of atoms $M \subseteq \at_C(T_{dc})$ is a model
if it (propositionally) satisfies the constraints in $T_C$ and if its
closure (propositionally) satisfies the post-constraints in $T_{PC}$. Thus,
the semantics of $\datac$ retains much of the simplicity of the semantics
of propositional logic.  $\datac$ was introduced by the authors in \cite{et00}.

\section{Applicability of the System} 

$\datac$ can be used as a computational tool to solve search problems.
We define a search problem $\Pi$ to be determined
by a set of finite {\em instances}, $D_\Pi$, such that for each instance
$I \in D_\Pi$, there is a finite set $S_\Pi(I)$ of all
{\em solutions} to $\Pi$ for the instance $I$. For example, the problem
of finding a hamilton cycle in a graph is a search problem: graphs are
instances and for each graph, its hamilton cycles (sets of their edges)
are solutions. 

A $\datac$ theory $T_{dc} = (T_C,T_H,T_{PC})$ {\em solves}
a search problem $\Pi$ if solutions to $\Pi$ can be computed (in
polynomial time) from answer sets to $T_{dc}$. Propositional logic and stable
logic programming \cite{mt99,nie98} are used as problem solving 
formalisms following the same general paradigm. 

We will illustrate the use of DC to solve search problems by discussing
the problem of finding a hamilton cycle in a directed graph.
Consider a directed graph $G$ with the vertex set $V$ and the edge set $E$.
Consider a set of atoms $\{hc(a,b)\colon (a,b)\in E\}$. An intuitive
interpretation of an atom $hc(a,b)$ is that the edge $(a,b)$ is in a
hamilton cycle. Include in $T_C$ all clauses of the form
$\neg hc(b,a)\vee \neg hc(c,a)$, where $a,b,c\in V$, $b\not=c$ and
$(b,a), (c,a) \in E$. In addition,
include in $T_C$ all clauses of the form $\neg hc(a,b)\vee \neg
hc(a,c)$, where $a,b,c\in V$, $b\not=c$ and $(a,b), (a,c) \in E$.
Clearly, the set of propositional variables of the form $\{hc(a,b)\colon
(a,b)\in F\}$, where $F\subseteq E$, satisfies all clauses in $T_C$ if
and only if no two
different edges in $F$ end in the same vertex and no two different edges
in $F$ start in the same vertex. In other words, $F$ spans a collection
of paths and cycles in $G$.

To guarantee that the edges in $F$ define a hamilton cycle, we must
enforce that all vertices of $G$ are reached by means of the edges in $F$
if we start in some (arbitrarily chosen) vertex of $G$. This can be
accomplished by means of a simple Horn program. Let us choose a vertex,
say $s$, in $G$. Include in $T_H$ the Horn rules $hc(s,t)\rightarrow
vstd(t)$, for every edge $(s,t)$ in $G$. In addition, include in
$T_H$ Horn rules $vstd(t),hc(t,u)\rightarrow vstd(u)$, for every edge
$(t,u)$ of $G$ not starting in $s$. Clearly, the least model of $F\cup
T_H$, where $F$ is a subset of $E$, contains precisely these
variables of the form $vstd(t)$ for which $t$ is reachable from $s$ by a
{\em nonempty} path spanned by the edges in $F$. Thus, $F$ is the set of
edges of a hamilton cycle of $G$ if and only if the least model of
$F\cup T_H$, contains variable $vstd(t)$ for every vertex $t$ of $G$.
Let us define $T_{PC} =\{vstd(t)\colon t\in V\}$ and
$T_{ham}(G)=(T_C,T_H,T_{PC})$. It follows that hamilton cycles of $G$
can be reconstructed (in linear time) from answer sets to the $\datac$
theory $T_{ham}(G)$. In other words, to find a hamilton cycle in $G$, it
is enough to find an answer set for $T_{ham}(G)$.

This example illustrates the simplicity of the semantics of $\datac$  --- it is only
a slight adaptation of the semantics of propositional logic to the case
when in addition to propositional clauses we also have Horn rules in
theories. It also illustrates the power of $\datac$ to generate concise
encodings. All known propositional encodings of the hamilton-cycle
problem require that additional variables are introduced to ``count''
how far from the starting vertex an edge is located. Consequently,
propositional
encodings are much larger and lead to inefficient solutions to the
problem.

\section{Description of  the System}

In this section we will discuss general features of
$\datac$. First, we will discuss the language
for encoding problems and give an example by showing the
encoding of the hamiltonicity problem.  Second, we will describe how we
execute the $\datac$ system. Third, we will give some details
concerning the implementation of {\tt dcs}, our solver.  
Last we discuss the expressitivity of $\datac$.

In the previous section we described the logic of $\datac$ in
the propositional case.  Our definitions can be extended to
the predicate case (without function symbols).  Each
constraint is treated as an abbreviation of a set of its
ground substitutions. This set is determined by the set of
constants appearing in the theory and by additional conditions
associated with the constraint ( we illustrate this idea later in 
this section).
When constructing predicate $DC$-based solutions to
a problem $\Pi$ we separate the representation of an instance to $\Pi$
from the constraints that define $\Pi$.  
The representation of an instance of $\Pi$ is a collection of facts or
the extensional database (EDB).
The constraints and rules that define $\Pi$ will be referred to as the intensional
database (IDB) and  the language for writing the problem descriptions that
constitute the IDB will be referred to as $L_{dc}$.
The separation of IDB and EDB means only one predicate description of
$\Pi$ is needed.  

The modules of $\datac$, {\tt ground} and {\tt dcs} provide a complete system for
describing and finding solutions to problems.  
An IDB in $L_{dc}$ along with a specific EDB
are the input to {\tt ground}. A grounded propositional 
theory $T_{dc}$ is output by {\tt ground} and used as input to {\tt dcs}.

A problem description in $L_{dc}$ defines predicates,
declares variables, and provides a description of the problem using rules.  
The predicates
are defined using types from the EDB. Similarly, the variables are 
declared using types from the EDB. The rules consist of constraints,
Horn rules and post-constraints. The constraints and post-constraints
use several constructs to allow a more natural modeling.  These
constructs could be directly translated to clauses. 
(We use them as shorthands to ensure the conciseness of encodings.)

We present here a brief discussion of the constraints, Horn rules 
and post-constraint in $L_{dc}$ and their meanings. 
Let $PRED$ be the set of predicates occurring in the IDB. 
For each variable $X$ declared in the IDB  the range $R(X)$ of $X$ 
is determined by the EDB. 

\begin{description}
	\item[Select$(n,m,\vec{Y};p_1(\vec{X}),\dots,p_i(\vec{X},\vec{Y})) q(\vec{X},\vec{Y})$,] 
		where $n,m$ are nonnegative integers such that $n \leq m, q \in PRED$ and
		$p_1, \dots ,p_i$ are  
		EDB predicates or logical conditions (logical conditions can be comparisons of
		arithmetic expressions or string comparisons). 
		The interpretation of this constraint is as follows:
		for every $\vec{x} \in R(\vec{X})$  
		at least $n$ atoms and at most $m$ atoms in the set
		$\{q(\vec{x},\vec{y}): \vec{y} \in R(\vec{Y}) \}$ are true. 

	\item[Select$(n,m,\vec{Y}) q(\vec{X},\vec{Y})$,]
		where $n,m$ are nonnegative integers such that $n \leq m, q \in PRED$.
		The interpretation of this constraint is as follows:
		for every $\vec{x} \in R(\vec{X})$  
		at least $n$ atoms and at most $m$ atoms in the set
		$\{q(\vec{x},\vec{y}): \vec{y} \in R(\vec{Y}) \}$ are true. 

	\item[Select$(n,m) q_1(\vec{X}), \dots, q_j(\vec{X})$,] 
		where $n,m$ are nonnegative integers such that $n \leq m, 
		q_1, \dots, q_j \in PRED$. 
		The interpretation of this constraint is as follows:
		for every $\vec{x} \in R(\vec{X})$  
		at least $n$ atoms and at most $m$ atoms in the set
		$\{q_1(\vec{x}), \dots, q_j(\vec{x}) \}$ are true. 
\end{description}
	Certain choices of $n,m$ in any of the Select constraints 
		allow construction of even more specific constraints in $T_{dc}$:\\
		$n=m$ exactly $n$ atoms must be true.\\ 
		$n=0$ at most $m$ atoms must be true.\\ 
		$m=999$ at least $n$ atoms must be true. 
\begin{description}
	\item[NOT $q_1(\vec{X}), \dots, q_i(\vec{X})$,]
		where $q_1, \dots, q_i\in PRED$.
		For every $\vec{x} \in R(\vec{X})$  
		at least one atom in the set
		$\{q_1(\vec{x}), \dots, q_i(\vec{x}) \}$ must be false.

	\item[$q_1(\vec{X})| \dots | q_i(\vec{X})$,]
		where $q_1, \dots, q_i\in PRED$.
		For every $\vec{x} \in R(\vec{X})$  
		at least one atom in the set
		$\{q_1(\vec{x}), \dots, q_i(\vec{x})\}$ must be true. 

	\item[$p_1(\vec{X}),\dots ,p_i(\vec{X})\rightarrow q_1(\vec{X})| \dots | q_j(\vec{X})$,]
		where $p_1, \dots, p_i, q_1, \dots, q_j\in PRED$.
		For every $\vec{x} \in R(\vec{X})$  if all atoms in the set 
		$\{ p_1(\vec{x}), \dots, p_i(\vec{x})\}$ are true
		then at least one atom in the set
		$\{ q_1(\vec{x}), \dots, q_j(\vec{x})\}$ must be true. (This constraint
		represents standard propositional logic clauses.)

	\item[$p_1(\vec{X}),\dots ,p_i(\vec{X})\rightarrow q_1(\vec{X}), \dots ,q_j(\vec{X})$,]
		where $p_1, \dots, p_i, q_1, \dots, q_j\in PRED$.
		For every $\vec{x} \in R(\vec{X})$  
		if all atoms in the set 
		$\{p_1(\vec{x}), \dots, p_i(\vec{x})\}$ are true
		then all atoms in the set
		$\{q_1(\vec{x}), \dots, q_j(\vec{x})\}$ must be true.

	\item[Horn $p_1(\vec{X}),\dots ,p_i(\vec{X})\rightarrow q_1(\vec{X}), \dots ,q_j(\vec{X})$,]
		where $p_1, \dots, p_i, q_1, \dots, q_j\in PRED$.
	\end{description}

\begin{figure}
\begin{center}
\begin{tabular}{|l|}
\hline
An example of a graph file used as EDB\\
\hline
vtx(1).\\
vtx(2).\\
vtx(3).\\
edge(1,3).\\
edge(3,2).\\
\hline
\end{tabular}
\end{center}
\caption{Format for EDB.}
\label{graph}
\end{figure}

\subsection{Methodology}
\label{method}


Here we show the encoding of a problem in $\datac$.  We will use the
example of hamiltonicity of a graph which we discussed previously.  
Figure \ref{graph}
shows the EDB format used by {\tt ground}.  This format is compatible 
to that used by {\tt smodels} and others.  
However, we {\em require} that the EDB be in separate files from the
IDB.  The format for the EDB allows data to be entered as sets, ranges, or
individual elements and constant values can be entered on the command line.

The IDB provides a definition of the problem in $L_{dc}$. The IDB file has three
parts. First a definition of the predicates, next the declaration of variables
and last a set of constraints and Horn clauses.
The types used in the IDB must be in the data file(s). For example, the only data
types in the graph file (Fig. \ref{graph}) are {\bf vtx} and {\bf edge}. Thus the
only data types which can be used in the IDB are {\bf vtx} and {\bf edge}.

\begin{figure}
\begin{center}
\begin{tabular}{|l|}
\hline
An example of a file used as IDB\\
\hline
\% comments begin with percent sign\\
idbpred \% section for defining predicates\\
vstd(vtx).\\
hc(vtx,vtx).\\
idbvar \% section for declaring variables\\
vtx X,Y.\\
idbrules \% rule section \\
\% constraints\\
Select(1,1,Y;edge(X,Y)) hc(X,Y).\\
Select(1,1,X;edge(X,Y)) hc(X,Y).\\
\% Horn rules\\
Horn Forall(X,Y;X!=i,edge(X,Y))\\ vstd(X), hc(X,Y) $\rightarrow$
vstd(Y).\\
Horn Forall(X,Y;X==i,edge(X,Y))\\ hc(X,Y) $\rightarrow$ vstd(Y).\\
\% post-constraints\\
vstd(X).\\
\hline
\end{tabular}
\end{center}
\caption{File showing IDB for the hamiltonicity of a graph. The types are from
Fig. \ref{graph}
\label{idb}}
\end{figure}

In the {\bf idbpred} section of Fig. \ref{idb}, we define two predicates, 
the {\em vstd} predicate
and the {\em hc} predicate.  The {\em vstd} predicate has one parameter
of type {\em vtx} and {\em hc} has two parameters both of type {\em vtx}.

The {\bf idbvar} section of Fig. \ref{idb}
declares two variables $X$, $Y$ both of type {\em vtx}.

The section containing the constraints, Horn clauses,
and post-constraints is proceeded by the keyword {\bf idbrules} 
(see Fig.\ref{idb}).
The order in which the rules  are entered is not 
important.  
The first constraint, $Select(1,1,Y;edge(X,Y)) hc(X,Y).$, ensures that each
vertex has exactly one outgoing edge.  The second constraint,
$Select(1,1,X;edge(X,Y)) hc(X,Y).$, requires that each vertex has exactly one
incoming edge.

The first Horn rule ranges over all $X,Y$ such that $edge(X,Y) \in EDB$ 
and $X \neq i$ where $i$ is a constant used to initialize $vstd(i)$. This
rule
requires both $vstd(X)$ and $hc(X,Y)$ to be true before we can assign the
value true to $vstd(Y)$.  The second Horn rule only requires $hc(X,Y)$
to be true before $vstd(Y)$ is assigned value true; however, in this
rule $X$ is restricted
to the value of $i$.  

The last line is a  post-constraint that requires $vstd(X)$ to be true
for all $X \in R(X)$ ensuring that the cycle is closed.

\subsection{Running the system}
\label{running}

Here we will describe the steps for execution of the $\datac$ system.
The first module of $\datac$, {\tt ground}, has as input a data file(s),
a rule file and command line arguments.  Output is the theory file which
will be used as input to {\tt dcs}. 
The theory is written to a file whose name is the catenation of the
constants and file names given as command line arguments. The extension {\bf .tdc}
is then appended to the output file name.
The command line arguments for {\tt ground}:\\

{\bf ground -r rf -d df [-c label=v] [-V]}\\

{\bf Required arguments}

\begin{itemize}
\item[-r]
{\bf rf} is the file describing the problem.  There must
be exactly one rule file.
\item[-d]
{\bf df} must be one or more
files containing data that will be used to instantiate the theory.
It is often convenient to use
more than one file for data.  
\end{itemize}

{\bf Optional arguments}

\begin{itemize}
\item[-c]
This option allows use of constants in both data and rule
files.  When {\bf label} is found while reading either file it is
replaced by {\bf v}.  {\bf v} can be any string that is valid
for the data type. If {\bf label} is to be used in a range then
{\bf v} must be an integer.  For example, if the data file contains
the entry {\bf queens[1..q].} then we can define the constant {\bf q} with
the option {\bf -c q=8}.  If more than one constant is needed then
{\bf -c b=3 n=14} defines both constants {\bf b,n} using the {\bf -c} option.
\item[-V]
The verbose options sends output to stdout
during the execution of {\bf ground}. This  output may
be useful for debugging of the data or rule files.  
\end{itemize}

For the example of hamiltonicity we could have a
data file (see Fig. \ref{graph}) named {\bf 1.gph} and an IDB file 
(see Fig. \ref{idb}) named
{\bf hcp}.  The constant {\bf i} is needed in the IDB file to initialize
the first vertex in the graph. The command line argument would be:

{\bf ground -r hcp -d 1.gph -c i=1}

The theory file produced would be named {\bf 1\_1.gph\_hcp.tdc}.

The second module of the $\datac$ system, {\tt dcs}, has as input
the theory file produced by {\tt ground}.
A file named dcs.stat is created or appended with statistics concerning
the results of executing {\tt dcs} on the theory. The command line
arguments are:\\

{\bf dcs -f filename  [-A] [-P] [-C] [-V]}\\

{\bf Required arguments}

\begin{itemize}
\item[-f]
{\bf filename} is the name of the file containing a theory produced by
{\tt ground}.
\end{itemize}

{\bf Optional arguments}

\begin{itemize}
\item[-A]
Prints the positive atoms for solved theories in readable form.  
\item[-P]
Prints the input theory and then exits.
\item[-C]
Counts the number of solutions.  This information is recorded in the statistics
file.
\item[-V]
Prints information during execution (branching, backtracking, etc).  Useful 
for debugging.
\end{itemize}

\subsection{Discussion of {\tt dcs}}
\label{dcs}

The $\datac$ solver uses a Davis-Putnam type approach, with
backtracking, propagation and LookAHead 
to deal with constraints represented as clauses, {\em select}
constraints and Horn rules, and to search for answer sets. The LookAHead
in $\datac$ is similar to local processing performed in {\tt csat}
\cite{dub96}.
However, we use different methods to determine how many literals to
consider in the LookAHead phase. Other techniques, especially 
propagation and search heuristics, were designed specifically for the
case of $\datac$ as they must take into account the presence of Horn
rules in programs.

Propagation consists of methods to reduce the theory. Literals which appear
in the heads of Horn rules, $l_h \in \at_H$ require different interpretations.  
A literal is a $l_h$ if and only if it appears in the head of a Horn rule.  
We can not guess an assignment for $l_h$ rather  it must be computed.  
We can only assign value true to
$l_h$ if it appears in the head of a Horn rule for which all literals in the
body of the Horn rule have been assigned the value true.
If one or more literals in the body of a Horn rule have been assigned the value
false then that rule is removed.  If  $l_h$ has not been assigned a value
and all Horn rules in which $l_h$ appeared in the head have been removed 
then $l_h$ is assigned the value false.  If we have a post-constraint that
required a value be assigned to $l_h$ and the value cannot be computed then
we must backtrack.

Non Horn rules are constraints which must be satisfied.  These
rules are identified by tags which indicate which method is needed to
evaluate the constraint during propagation.

The LookAHead procedure tests a number of literals not yet assigned a value.
The LookAHead procedure is similar to the local processing procedure used for
{\tt csat} \cite{dub96}.
A literal is chosen, assigned the value true, then false using propagation 
to reduce and evaluate the resulting theories.
If both evaluations of assignments result in conflicts then we return
false and backtracking will result. If only one evaluation results in conflict
then we can assign the literal the opposite value and continue the LookAHead
procedure.  If neither
evaluation results in conflict we cannot make any assignments, but we save
information (number of forced literals and satisfied constraints) computed
during the evaluations of the literal.  

The number of literals to be tested has been determined empirically.  It is
obvious that if all unassigned literals were tested during each LookAHead it
would greatly increase the time.  However, if only a small number of literals are to be
tested during each LookAHead then they must be chosen to provide the best
chance of reducing the theory.  To choose the literals with the best chance
of reducing the theory, we order the unassigned literals based on a sum computed
by totaling the weights of the unsatisfied constraints in which they appear.  The 
constraint weight is 
based on both the length and type of constraint. The shorter the constraint the 
larger the weight and when literals are removed the weight is recomputed.
The literals are tested in descending order of the sum of constraints weights.
Using this method we need to test only a very small number
of literals during each LookAHead to obtain good results.

At the completion of the LookAHead procedure, we use the information 
computed during the evaluation of literals to
choose the next branching literal and its initial truth assignment.

\subsection{Expressitivity}
\label{express}

The expressive power of $\datac$ is the same as that of logic
programming with the stable-model semantics.  The following
theorem is presented in \cite{et00}

\begin{theorem}
The expressive power of $\datac$ is the same as that of stable logic
programming. In particular, a decision problem $\Pi$ can be solved
uniformly in $\datac$ if and only if $\Pi$ is in the class {\rm NP}.
\end{theorem}


\section{Evaluating the System}

The $\datac$ system provides a language, $L_{dc}$, which facilitates
writing problem descriptions.  Once an IDB is written in $L_{dc}$
it can be used for any instance of the problem for which data,
in the EDB format, is available or can be generated.  
It is possible to 
add constraints to IDB for a given problem 
when new requirements or constraints occur.
The constructs used in $L_{dc}$ allow
for a natural description of constraints.
Users need only know enough about a specific problem to be able to
describe the problem in $L_{dc}$ (there is a user's manual with
examples). To help with programming in $\datac$
{\tt ground} provides error messages and compiling 
information that are useful for debugging the IDB.

\begin{table}
\begin{center}
\begin{tabular}{|r|r|r|r|}
\hline
\multicolumn{1}{|c|}{B-N} &
\multicolumn{1}{|c|}{csat} &
\multicolumn{1}{|c|}{dcs} &
\multicolumn{1}{|c|}{smodel} \\
\hline
b-n&sec&sec&sec\\
\hline
3-13 &  0.03 & 0.00 & 0.12\\
3-14 &  0.05 & 0.00 & 0.16 \\
4-14 &  0.05 & 0.01 &  0.23 \\
4-43 &  0.59 & 1.91 & 5.23 \\
4-44 &  1.95 & 51.04 & 5.55 \\
4-45 &  1599.92 & 226.44  & 12501.00 \\
\hline \end{tabular} \end{center}

\caption{Schur problem; times}
\label{schur}
\end{table}

\subsection{Benchmarks}

The $\datac$ system has been executed using problems from NP,
combinatorics, and planning. In particular, it has been used
to compute hamilton cycles and colorings in graphs, to solve the
$N$-queens problem, to prove that the pigeonhole problem has no solution
if the number of pigeons exceeds the number of holes, and to compute
Schur numbers.  

\begin{figure}
\centerline{\hbox{\psfig{figure=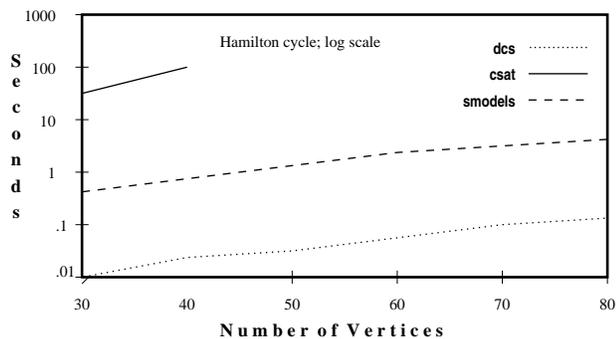}}}
\caption{Results of computing hamilton cycles;log scale}
\label{ham}
\end{figure}

The instances for computing hamilton cycles were obtained by randomly generating one thousand
directed graphs with $v=30,40 \ldots , 80$ and density such that $ \approx 50\%$
of the graphs contained hamilton cycles.

\begin{figure}
\centerline{\hbox{\psfig{figure=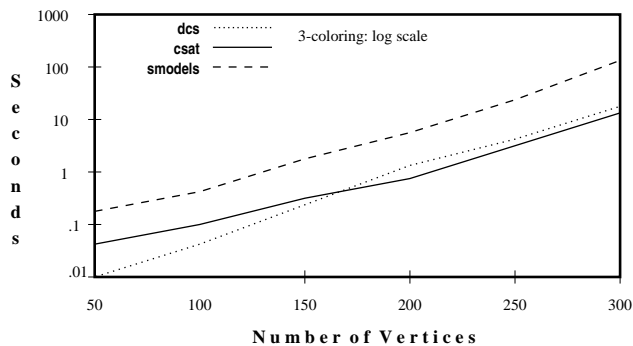}}}
\caption{Results on encodings of coloring problems;log scale}
\label{color}
\end{figure}

The graphs for instances of coloring were one hundred randomly generated graphs
for $v=50,100, \ldots , 300$ with density such that $\approx 50\%$ had solutions 
when encode as 3-coloring.

\subsection{Comparison} 

We have used the benchmarks to compare $\datac$ with systems based
on stable model semantics and satisfiability.
The performance of $\datac$ solver {\tt dcs} was compared with {\tt smodels},
a system for computing stable models of logic programs \cite{ns96}, and
{\tt csat}, a systems for testing propositional 
satisfiability \cite{dub96}.
In the case of {\tt smodels} we used version 2.24 in conjunction with
the grounder {\tt lparse}, version 0.99.41 \cite{nie98}.
The extended rules \cite{sim99} allowed in the newer 
versions of {\tt smodels} and {\tt lparse} were used where applicable.
The programs were all executed on a Sun SPARC Station
20. For each test we report the cpu user times for processing the
corresponding propositional program or theory. We tested all three system
using the benchmarks discussed in the previous section.

\begin{figure}
\centerline{\hbox{\psfig{figure=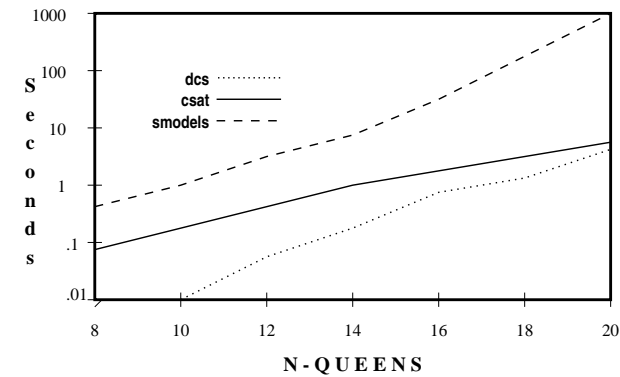}}}
\caption{N-queens problem;log scale}
\label{queen}
\end{figure}

\begin{figure}
\centerline{\hbox{\psfig{figure=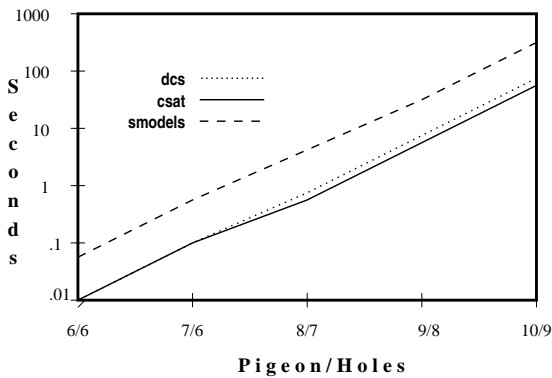}}}
\caption{Pigeonhole problem ;log scale}
\label{ph}
\end{figure}

\begin{table}
\begin{tabular}{|l||r|r||r|r|}
\hline
Theory &Vertices &Edges & Atoms & Clauses \\
\hline
DC &30 &130 & 160 & 220 \\
smodels &30 &130 & 1212 & 621\\
SAT& 30 &130 & 900 & 27960\\
\hline \end{tabular} 
\caption{Difference in size of theories for hamilton cycle}
\label{tsize}
\end{table}

Note that {\tt csat} performs comparable to $\datac$ for pigeonhole (see Fig.
\ref{ph}), N-queens (see Fig. \ref{queen})
and coloring (see Fig. \ref{color}).  These are problems 
where $\datac$ encodings do not use Horn rules 
(in the examples here only encodings for computing hamilton cycles use Horn rules). 
The closure properties of Horn rules allow for much smaller theories as shown in
Table \ref{tsize}.
The satisfiability theories for computing hamilton
cycles are so large that they were not practical to execute for over $40$ vertices
(see Fig. \ref{ham}). {\tt smodels} performs much better than satisfiability
solvers for computing the hamilton cycles although not as well as $\datac$. 
The results for computing Schur numbers (see Table \ref{schur}) also show
much better results for $\datac$.

Experimental results show that {\tt dcs} often outperforms systems based
on satisfiability as well as systems based on non-monotonic logics, and
that it constitutes a viable approach to solving problems in AI, constraint
satisfaction and combinatorial optimization.  We believe that our focus on 
short encodings (see Fig. \ref{tsize}) is the key to the success of {\tt dcs}.



\end{document}